\documentclass[letterpaper, 10 pt, conference]{ieeeconf}

\usepackage{amsthm}
\usepackage{amsmath}
\usepackage{amssymb}
\usepackage{graphicx}

\usepackage{color}

\newtheorem{thm}{\protect\theoremname}

\newtheorem{prop}[thm]{\protect\propositionname}
\newtheorem{rem}[thm]{\protect\remarkname}

\providecommand{\definitionname}{Definition}
\providecommand{\propositionname}{Proposition}
\providecommand{\remarkname}{Remark}
\providecommand{\theoremname}{Theorem}
\providecommand{\lemmaname}{Lemma}
\providecommand{\corollaryname}{Corollary}

\begin{document}

\title{Convex Model Predictive Control for Vehicular Systems}

\author{Tiffany A. Huang, Matanya B. Horowitz, Joel W. Burdick}

\maketitle
\thispagestyle{empty}
\pagestyle{empty}

\begin{abstract}
In this work, we present a method to perform Model Predictive Control (MPC) over systems whose state is an element of $SO(n)$ for $n=2,3$. This is done without charts or any local linearization, and instead is performed by operating over the orbitope of rotation matrices. This results in a novel MPC scheme without the drawbacks associated with conventional linearization techniques. Instead, second order cone- or semidefinite-constraints on state variables are the only requirement beyond those of a QP-scheme typical for MPC of linear systems. Of particular emphasis is the application to aeronautical and vehicular systems,  wherein the method removes many of the transcendental trigonometric  terms associated with these systems' state space equations. Furthermore,  the method is shown to be compatible with many existing variants of  MPC, including obstacle avoidance via Mixed Integer Linear Programming  (MILP).
\end{abstract}

\section{Introduction}

Model Predictive Control (MPC) has proven to be an invaluable tool
in the generation of trajectories for linear systems. Via linearization,
this method has been extended to nonlinear  \cite{todorov2005generalized, hauser2006barrier} systems, and those that evolve on Lie Groups \cite{saccon2011optimal},  as well. Unfortunately,
the linearization process introduces several difficulties, including
not only approximation error in the dynamics, but also computational
costs. These obstacles are particularly acute for aeronautical systems,
where the trigonometric terms present in the dynamics prevent the
use of linear methods directly. This paper proposes a new approach that does 
not require linearization to produce a solution, instead relying on a particular convex relaxation, thereby having the 
potential to reduce costs and obtain a globally optimal solution when applied to motion planning. 

The algorithm presented in this work relies upon the convex hulls of $SE(2)$ 
and $SE(3)$, which can be represented as linear matrix inequalities (LMI). 
LMIs are a class of convex constraints that have had a profound effect 
on the field of optimization and control in recent decades \cite{Boyd:1997uz, Boyd:2004uz}. 
Using the theory behind the new convex relaxation method, these
tractable convex relaxations allow a large number of robotic and vehicle motion 
planning problems to be handled using convex optimization with readily 
available commercial solvers. A convex optimization approach is significant, as it guarantees solutions 
are both rapidly computable and free of local minima. 

A faster and more efficient motion planning technique would have
several important applications regarding aeronautical systems. 
Spacecraft are particularly sensitive to fuel use. Similarly, 
the landing and collection of UAVs after missions is an important 
problem, as current UAV landing techniques are limited due to a
lack of real time autonomy. By allowing for dynamically feasible
trajectories to be calculated rapidly, reliability and rapid response
may be improved. Specifically, we examine a particular recapture
problem requiring a UAV to use a hook to latch onto a rope. The 
UAV will generate a trajectory to fly into the rope, latch on, and slide 
down for collection. This recapture method is currently widely 
used in defense and aerospace industries as well as in the U.S. 
Navy. We also demonstrate the generality of the method by applying it to the Dubin's car model, as well as a full, three dimensional satellite maneuvering problem.

\subsection{Related Work}
Trajectory planning for vehicular systems has long been an active research subject, with approaches primarily falling to two paradigms. The first tackles the problem through a sampling based approach \cite{frazzoli2001randomized, lavalle2006planning}, in which the reachable portion of the state space is grown via sampling, either from neighboring portions of the already reachable state space and/or from the input space.

An alternative paradigm lies in optimization, primarily based on a Model Predictive Control (MPC) framework, in which a model of the system is used to optimize a time-discretized trajectory into a finite horizon of the future.  References \cite{richards2002aircraft} and \cite{bellingham2002receding} shares many similarities with our work, adopting a Mixed Integer framework to allow for obstacles, and can be considered an antecedent of this work. Where our work differs is in our use of the convex hull of $SE(2)$, and $SE(3)$ to constrain the motion of the system, which had previously required an infinite number of Linear Program constraints and had thus only been approximated. Additionally, we develop the method further, incorporating integrator dynamics for our UAV and spacecraft examples.  The work of \cite{blackmore2012lossless} also uses convex programming for model predictive control. In particular, a norm constraint on the control thrust is relaxed, with the relaxation shown to be tight.  Going further in examining the Lie group structure of these systems, \cite{saccon2012constrained} develops a functional approach to the problem that circumvents the need for time discretization and grapples with the manifold structure of the group directly. 

Research into the structure and application of the convex hulls of $SE(n)$ has increased recently, a body of literature that the current paper builds upon. In \cite{orbitopes}, the authors study the structure of convex bodies of Lie groups broadly. In \cite{saunderson2014semidefinite} the structure of the convex hull of $SE(n)$ is studied in detail, yielding semidefinite descriptions of the convex hull for arbitrary $n$, and also uses these parameterizations to solve Wahba's problem, a common estimation problem in aeronautical navigation. In computer vision, the authors of \cite{horowitz2014convex} investigated the convex relaxation is pose estimation problems, integrating convex penalties to create novel convex optimization problems. The techniques were also integrated with decentralization schemes in \cite{matni:so_admm} to create a distributed algorithm for consensus over $SO(n)$.

\section{Background}
The focus of this work is in the reduction of planning over systems with states in the special euclidean group to a convex program. Convex programming via interior point methods is well studied  \cite{Boyd:2004uz}, and may produce fast results in 
practice. By definition, convex programming also guarantees that all solutions 
are global minima, eliminating the local minima problem encountered 
through linearized model predictive control. 

\subsection{Model Predictive Control}
Model predictive control is a framework in which
control inputs are solved for using an optimization problem that minimizes a certain 
cost function with the current state of the system as an input \cite{garcia1989model}. For a linear, discrete time system
\begin{align}
\label{eq:linear_mpc}
s(k + 1) &= As(k) + Bu(k) \nonumber \\
y(k) &= Cs(k),
\end{align}
where at time $k$, $s(k)$ is the state of the system, $u(k)$ is the input, 
and $y(k)$ is the regulated output. The goal is to minimize a cost function
\begin{equation}
\label{eq:J_cost}
J(\bar{s},\bar{u})  = \| s(T)-s_f \| _M + \sum_{k=1}^{T-1} \ell(s(k),u(k)),
\end{equation}
where the stage cost accrued at time $k$ is given by the function
\begin{equation}
\label{eq:L_inst_cost}
\ell(s,u) = \| s(k)-s_d(k)\|_Q + \|u(k)\|_R,
\end{equation}
where $\bar{s}$, $\bar{u}$ denote the system trajectories, $s_d$ a desired state, and $||\cdot||_Q$ the $Q$ weighted inner product. Here, $Q$ represents a state penalty that must be positive semidefinite, $R$ the control matrix that is restricted to positive definite, and $M$ a terminal semidefinite penalty. 

The problem with MPC for linear systems is that it is readily seen to be a Quadratic Program (QP) . However, many engineering problems, including those of most aeronautical systems are nonlinear. In these cases, most typical MPC schemes linearize the 
non-linear system at each time sample. 

Oftentimes, MPC iterates over a finite time horizon to generate an optimal trajectory. It executes one step of the  optimization and repeats the planning calculation, a process known as Receding Horizon Control (RHC).  A particular advantage of this method is its ability to constantly readjust to changes in the environment that might affect the generated trajectory. However, 
the need to start optimizing at each time step slows down 
the planning process considerably. Although MPC works well on many different 
and complex systems, linearization for complex systems may require significant computation and approximation. With the addition of many joints and robotic arms on spacecraft or with more 
complicated aerial vehicles like a quadrotor, the motion planning problem 
can be sufficiently complex to prevent the use of MPC for highly dynamic systems. Furthermore, due to the 
nature of the method, MPC does not guarantee a global minimum solution, only achieving a local minimum due to its local linearization.

\subsection{Orbitopes of $SO(2)$}
In this work, the systems analyzed are assumed to have state elements in $SE(2)$ and 
$SE(3)$. We begin by examining the $SE(2)$ constraint, and demonstrating how it may be relaxed to a spectrahedral constraint (i.e. representable as a convex constraint), before explaining how the framework extends to $SE(3)$. 

A matrix  containing the rotation $(R)$ and translation $(T)$ of the system at time $t$ 
is used to represent the state of the system in homogeneous coordinates. In $SE(2)$, the state of the 
system has $T=(x, y)^T$ coordinates representing the translation of the system 
and $R \in SO(n)$ representing the orientation of the system
\begin{align}
D &= \begin{bmatrix}
	R & T \\
	0 & 1
	\end{bmatrix},
\end{align}
where $R \in SO(n)$ and
\begin{equation}
SO(n) = \left\{ R \mid R\cdot R^T = I, \det(R)=+1\right\}.
\end{equation}
A parameterization over a rotation angle $\theta$ is possible in two dimensions through the use of trigonometric terms
\begin{equation}
	R =  \begin{bmatrix}
	\cos (\theta) & -\sin (\theta) \\
	\sin (\theta) & \cos (\theta) \\
	\end{bmatrix}.
\end{equation}

The new method described in this paper relies on the use of an \textit{orbitope} \cite{orbitopes}, 
the convex set consisting of convex hulls of the orbits given by a compact, algebraic group \textit{G} acting linearly on a real vector space. The convex hull conv $S$ is the set of all convex combinations of points in $S$. More specifically, this method deals with a group acting on its identity element, or the \textit{tautological orbitope}. It relaxes the constraint on $SO$(2), a rotation in 2 dimensions, from a unit sphere, $x^2 + y^2 = 1$, to a unit disk, $x^2 + y^2 \le 1$ by taking its convex hull, where $x = \text{cos } \theta$ and $y = \text{sin } \theta$. Through this relaxation, the authors have shown that the tautological orbitopes for \textit{SO(n)}, $n = 2, 3$ can be written as a linear matrix inequality (LMI) 
\cite{horowitz2014convex}. In two dimensions this is the following inequality,
\begin{align}
\text{conv}(SO(2)) &= \{ \begin{bmatrix}
				   a & b \\
				   -b & a
				   \end{bmatrix} : a^2 + b^2 \le 1 \} \nonumber \\
&= \{ \begin{bmatrix}
	a & b \\
	-b & a
	\end{bmatrix} : 
	\begin{bmatrix}
	I & \left( \begin{array}{c} a\\ b \end{array} \right) \\
	\begin{matrix}\begin{pmatrix}a & b\end{pmatrix}\end{matrix} & 1
	\end{bmatrix} \succeq 0 \}	.
\end{align}

As LMIs are convex constraints, we have relaxed the nonlinear optimization problem to a 
convex semidefinite problem, and in the special case of $SO(2)$, a second-order cone (SOC) 
problem \cite{Boyd:1997uz, Boyd:2004uz}. Thus, if a motion planning problem in $SO(n)$ has a convex objective, the LMI constraints representing the tautological orbitope of $SO(n)$ can 
be applied as convex restraints to transform the problem into a semidefinite 
optimization problem. 

\begin{rem}
The second-order cone contains additional structure over the semidefinite cone, structure that can be leveraged in computation. An example is in CVXGEN \cite{mattingley2012cvxgen}, which allows for embedded C-code to be generated for particular problem instances. This has allowed for problems with thousands of variables to be solved in milliseconds. Combined with other structural properties of MPC \cite{wang2010fast}, there are many opportunities for optimizing computation.
\end{rem}

\subsection{Orbitopes of $SO(n)$ for $n=3$}
\begin{figure*}[t!]
\begin{centering}
\begin{equation}
\left(\begin{array}{cccc}
1+x_{11}+x_{22}+x_{33} & x_{32}-x_{23} & x_{13}-x_{31} & x_{21}-x_{12}\\
* & 1+x_{11}-x_{22}-x_{33} & x_{21}+x_{12} & x_{13}+x_{31}\\
* & * & 1-x_{11}+x_{22}-x_{33} & x_{32}+x_{23}\\
* & * & * & 1-x_{11}-x_{22}+x_{33}
\end{array}\right)\succeq0\label{eq:sdp_rep_so3}
\end{equation}
\par\end{centering}
\caption{Spectrahedral representation of $\text{conv}(SO(3))$ \cite{orbitopes}.  Omitted $*$ elements indicate the symmetric completion of the matrix.}
\end{figure*}

The development of the orbitope for $SO(3)$ is more involved than that for $SO(2)$, and is presented in \cite{orbitopes, horowitz2014convex}. In brief, an explicit parameterization of $SO(3)$ is given by its embedding into the space of pure quaternions (a subgroup of $SU(2)$). Several transformations of the constraints that define this set allow for them to be placed in a form amenable to the following convex relaxation.

\begin{prop}
The tautological orbitope $\text{conv}\left(SO(3)\right)$ is a spectrahedron
whose boundary is a quartic hypersurface. In fact, a $3\times3$-matrix
$X=\left(x_{ij}\right)$ lies in $\text{conv}\left(SO(3)\right)$
if and only if it satisfies (\ref{eq:sdp_rep_so3}).
\end{prop}

\section{Convex Model Predictive Control over $SO(n)$}
We are concerned with the finite horizon control of systems which have a set of states $R(t)$ that are restricted to lie in the special orthogonal group, i.e. $R(t)\in SO(n)$. We will furthermore assume that the dynamics of the system are linear with respect to this variable. For the remainder of this work, we will in particular consider systems with such a state $R(t)$ that is the body orientation, and forward velocity or acceleration in this body frame. Specifically, we use models of the form
\begin{eqnarray}
\dot{R}(t) & = &R(t) + u(t) \\
\dot{s}(t) & = &R(t) V
\end{eqnarray}
where $V$ is a fixed forward velocity vector in the body frame, $s(t)$ is the systems cartesian state at time $t$, and $h$ is the time step. As will be seen, this class of problems encompasses a number of robotic systems, from the Dubins car to satellite systems.

We discretize such systems in time, and place them in the MPC framework, yielding optimizations of the following form.
\begin{eqnarray}
\min. &  & \sum_{t=1}^{T}l\left(s\left(t\right),u\left(t\right)\right)+\phi\left(x\left(T\right)\right)\\
s.t. &  & R(t+1)=R(t)+h u(t)\\
      &  & s(t+1)=s(t) + h R(t)\cdot V \label{eq:mpc_sen_cons}\\
      &  & R(t)\in SO(n), 
\end{eqnarray}
where $n=2,3$, and $V$ is some forward vector in the body frame. With the model problem formed, we will subsequently refer to $SE(n)$, with only the cases $n=2,3$ in mind. 

Via orbitopes, equation \eqref{eq:mpc_sen_cons} is replaced with the convex constraint $R(t)\in \text{Co}\left(SE\left(n\right) \right)$. The question then arises: under what scenarios will this relaxation be exact, i.e. when will the solution to the relaxed and original problems coincide? In related work \cite{horowitz2014convex}, it was demonstrated that such a guarantee may be provided for a wide variety of computer vision problems. Unfortunately, no such guarantee is available in the MPC context. Instead, we propose optional non-convex restrictions on the variable $R(t)$ based on Mixed-Integer constraints. Including these restrictions will allow us to solve many motion planning problems that are by nature non-convex. 

Systems with second-order dynamics on the rotation component $R(t)$ are handled in a similar fashion, with the simple addition of a state variable, for instance a variable $\dot{R}(t) = S(t)$, where $\dot{S}(t)=u(t)$. In this case, the constraint applies only to $R(t)\in \text{conv}(SO(n))$ and ensures that only valid inputs are created.

\section{Mixed Integer Linear Programming}
In the context of MPC, it has been found that Mixed Integer constraints
may be useful in encoding non-convex constraints in the domain \cite{richards2002aircraft}.
Our incorporation of Mixed Integer Linear Programs (MILPs) into the 
framework is twofold: the first is to demonstrate that existing MPC 
methodologies apply readily; and the second is to 
constrain the solutions within the convex hull of $SO(n)$ in order to incorporate additional control design criteria. For 
aeronautical applications, this allows the enforcement of dynamic 
constraints such as minimum speed.

Mixed integer linear programming is a modification of a linear
program (LP) in which some of the constraints take only integer 
values. For our purposes, the constraints will only take 
binary values, 0 or 1. MILP allows for the inclusion of discrete 
decisions, which are non-convex in the optimization problem. For 
instance, integer constraints enable capabilities such as obstacle 
avoidance in which a vehicle must determine whether to go "left" 
or "right" around an obstacle. In general, MILPs are NP-complete \cite{floudas1995nonlinear}, but in practice suboptimal solutions may be calculated quickly, and globally optimal solutions are typically found in a reasonable amount of time for control problems \cite{richards2005mixed}. 

We review the encoding of obstacle avoidance into MILP constraints, following \cite{richards2005mixed}. For the two dimensional case with a 
rectangular obstacle, $(x_\text{min}, y_\text{min})$ is defined as the 
lower left-hand corner of the obstacle and $(x_\text{max}, y_\text{max})$ 
is defined as the upper right-hand corner of the obstacle. Points on the 
path of the vehicle $(x,y)$ must lie outside of the obstacle, which can be 
formulated as the set of rectangular constraints.

To convert these conditions to mixed integer form, we introduce
an arbitrary positive number $M$ that is larger than any
other variable in the problem. We also introduce integer
variables $a_k$ that take values 0 or 1. The rectangular constraints 
now take the form
\begin{align}
\label{eq:milp_cons}
x &\le  x_\text{min} + M a_1,  \\
-x &\le  -x_\text{max} + M a_2, \nonumber \\
y &\le  y_\text{min} + M a_3, \nonumber \\
-y &\le  -y_\text{max} + M a_4, \nonumber \\
\sum\limits_{k=1}^4 a_k &\le  3. \nonumber
\end{align}
Note that if $a_k$ = 1, then the $k^\text{th}$ constraint from (\ref{eq:milp_cons})
is relaxed. If $a_k$ = 0, then the $k^\text{th}$ constraint is
enforced. The final MILP constraint ensures that at least one
condition from (\ref{eq:milp_cons}) is enforced and the vehicle will avoid the
obstacle.


\subsection{Convex Hull MILP Constraints}
To use MILP to constrain solutions within the convex hull of $SO(2)$,
an "obstacle" is introduced into the convex hull of $SO(2)$. The hull of $SO(2)$, as has been shown,
can be represented as a circle of radius 1 centered at the origin, and the obstacle corresponds to a region within the circle that is admissible, preventing the determinant from obtaining a set of values. This is of importance as the forward velocity vector in the world frame is dependent on the forward velocity in the body frame, $M$ in (\ref{eq:mpc_sen_cons}), rotated by $R$. Thus, a decrease in the determinant of $R$ corresponds to a lower velocity, and a slowing of the vehicle. In this approach, the determinant of $R$ can be less than 1 because we relaxed the constraints on $SO(n)$ to its convex hull. While appropriate in some contexts, e.g. vehicles, it is undesirable in others, such as an aircraft.

\begin{figure}
\begin{centering}
\includegraphics[scale=0.33]{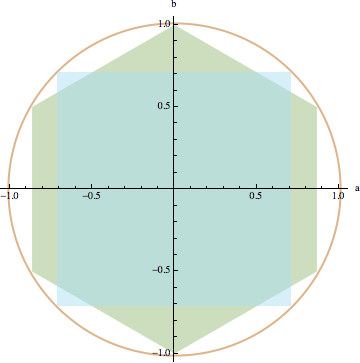}
\includegraphics[scale=0.33]{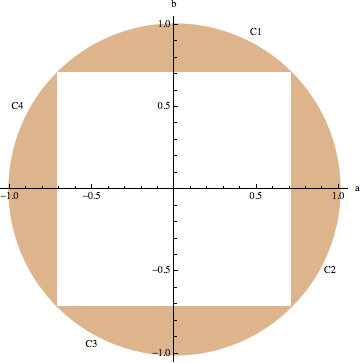}
\end{centering}
\caption{Mixed-Integer constraints applied to the determinant of $conv(SO(2))$. On the left, we illustrate square or hexagonal regions that are inadmissable for the variables in the rotation $R \in conv(SO(2))$. The resulting admissible regions are a union of individually convex sets, illustrated in the right as $C_1,\ldots,C_4$ for the square region.
\label{fig:viz_mi_cons}}
\end{figure}

Ideally, this obstacle would be a circle that restricts the determinant of R from
reaching 0. However, MILP constraints cannot represent a circular
obstacle, so we start with a square obstacle by using the same form of
constraints shown in (\ref{eq:milp_cons}). An example of the determinant constraint looks like
\begin{align}
\label{eq:det_cons}
sx_\text{min} &= -\sqrt{2}/2 \\
sx_\text{max} &= \sqrt{2}/2 \nonumber \\
sy_\text{min} &= -\sqrt{2}/2 \nonumber \\
sy_\text{max} &= -sqrt{2}/2 \nonumber \\
a &\le  sx_\text{min} + M c_1,  \nonumber \\
-a &\le  -sx_\text{max} + M c_2, \nonumber \\
b &\le  sy_\text{min} + M c_3, \nonumber \\
-b &\le  -sy_\text{max} + M c_4, \nonumber \\
\sum\limits_{k=1}^4 c_k &\le  3. \nonumber
\end{align},
where $a$ and $b$ are elements of the rotation matrix $R \in SO(2)$:
\begin{equation}
R = \begin{bmatrix}
	a & -b \\
	b & a 
	\end{bmatrix}
\end{equation}
Illustrations of the determinant constraint are shown in Figure \ref{fig:viz_mi_cons}. As the number of faces is increased, the symmetry of the minimum speed constraint may be improved as desired.

%

\subsection{Projection onto the Convex Hull of $SO(n)$}
In the case where the optimal solution is not an element of $SE(n)$, a practical algorithm will require a method of quickly generating a ``nearby'' Euclidean transformation.  As a potential heuristic, we propose using a projection of the inadmissible transformation onto the admissible set.  When this projection is taken with respect to the Frobenius norm, a simple singular value decomposition (SVD) based solution exists \cite{Belta:2002euclidean}.

In particular, let a rotation $S \in \text{conv}(SO(n))$ have the SVD
\begin{equation}
S = U \Sigma V^T.
\end{equation}
Then $S' = U V^T$ is the projection of $S$ onto $SO(n)$, i.e. $S' = \arg \min \{ \|T-S\|_\text{F} : T \in SO(n)\}.$  Future work will look to quantify the performance of such heuristics.

\section{Examples}
We will investigate three examples: the Dubin's car, UAV retrieval, and 
spacecraft planning.  The MATLAB package CVX \cite{cvx} was used to solve the convex optimization 
problems formulated for motion planning simulations.  MOSEK \cite{andersen2000mosek} was used as the numerical solver due to our need for Mixed Integer support.

\subsection{Dubin's Car}

\begin{figure}
\begin{centering}
\includegraphics[scale=0.45]{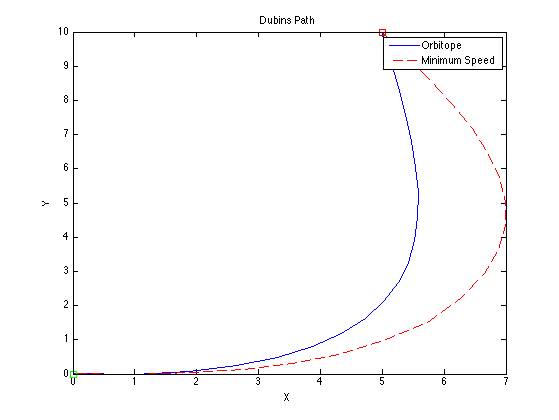}
\end{centering}
\caption{Example trajectories for the Dubins' car with and without a minimum speed of $det(R)=\frac{1}{2}$, corresponding to the right of Figure \ref{fig:viz_mi_cons}. The trajectory begins at the origin and ends at $(x,y)=(5,10)$.
\label{fig:dubins_traj}}
\end{figure}

\begin{figure}
\begin{centering}
\includegraphics[scale=0.45]{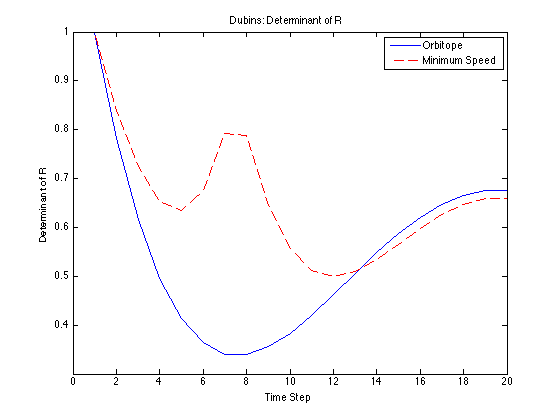}
\end{centering}
\caption{Determinant of the $SO(2)$ state $R$ for the Dubins' car trajectories shown in Figure \ref{fig:dubins_traj} with and without the determinant constraint.
\label{fig:dubins_r}}
\end{figure}


The Dubin's car example provides a simple framework
that can be extended to various types of vehicles and aeronautical systems.
In these examples, a velocity matrix $V$ was introduced to move the
vehicle forward a unit length in the $x$-direction with no rotation. Specifically, in (\ref{eq:mpc_sen_cons}), we have
\begin{equation}
V =  \left( \begin{array}{c} 1\\ 0 \end{array} \right) 
\end{equation}
and our translational state is $s=(x,y)^T$. 

For a vehicle following Dubin's model, the maximum turn 
rate results in a minimum turning radius, and backward motion is 
not allowed. The length of the path is optimized using the function
\begin{equation}
L(\vec{q},\vec{u}) = \int_0^{T} \sqrt{\dot{x}(t)^2 + \dot{y}(t)^2} \,dt
\end{equation}
where $T$ is the time when the goal is reached, $\dot{x}$ is the
velocity in the $x$-direction, and $\dot{y}$ is the velocity in the $y$-
direction. The configuration of the vehicle is designated by $q = (x, y, \theta)$.
The governing dynamics of this Dubin's car problem \cite{dubins1957curves} are given by
\begin{align}
\dot{x} &= z \text{ cos } \theta \nonumber \\
\dot{y} &= z \text{ sin } \theta \nonumber \\
\dot{\theta} &= u
\end{align}
where $z$ is the constant speed of the vehicle and $u$ is the turn
rate. Our example consisted of a vehicle with no inertia
attempting to reach a particular point in a time horizon of 20 seconds, discretized with time step $h=1.0$.
The final position of the vehicle was constrained to equal the given
goal point at time $t = T$. The optimal trajectory was generated, and
the determinant of the rotation matrix $R(k)$ was examined over 
time, with the results shown in Figures \ref{fig:dubins_traj} and \ref{fig:dubins_r}.

In examining the determinant of the rotation matrix $R(k)$ we see that the convex relaxation is not always exact. When the solution is exact, the determinant of $R$ should
equal unity. Lower values of the determinant correspond to rotation matrices that lie inside the convex hull of $SO(2)$, with the interpretation being that the system is traveling at a lower velocity $v < z$. Indeed, in this context the ability to travel inside the convex hull is desirable. However, we also demonstrate the effect of imposing a minimum size on the determinant of $R$, also shown in the figures. With the minimum speed MILP constraints the solution took approximately one minute to compute on a 2.3Ghz Intel i7 processor, while without the mixed integer constraints it required approximately $0.1s$.

\subsection{UAV Retrieval}

\begin{figure}
\begin{centering}
\begin{tabular}{c c c}
\includegraphics[scale=0.13]{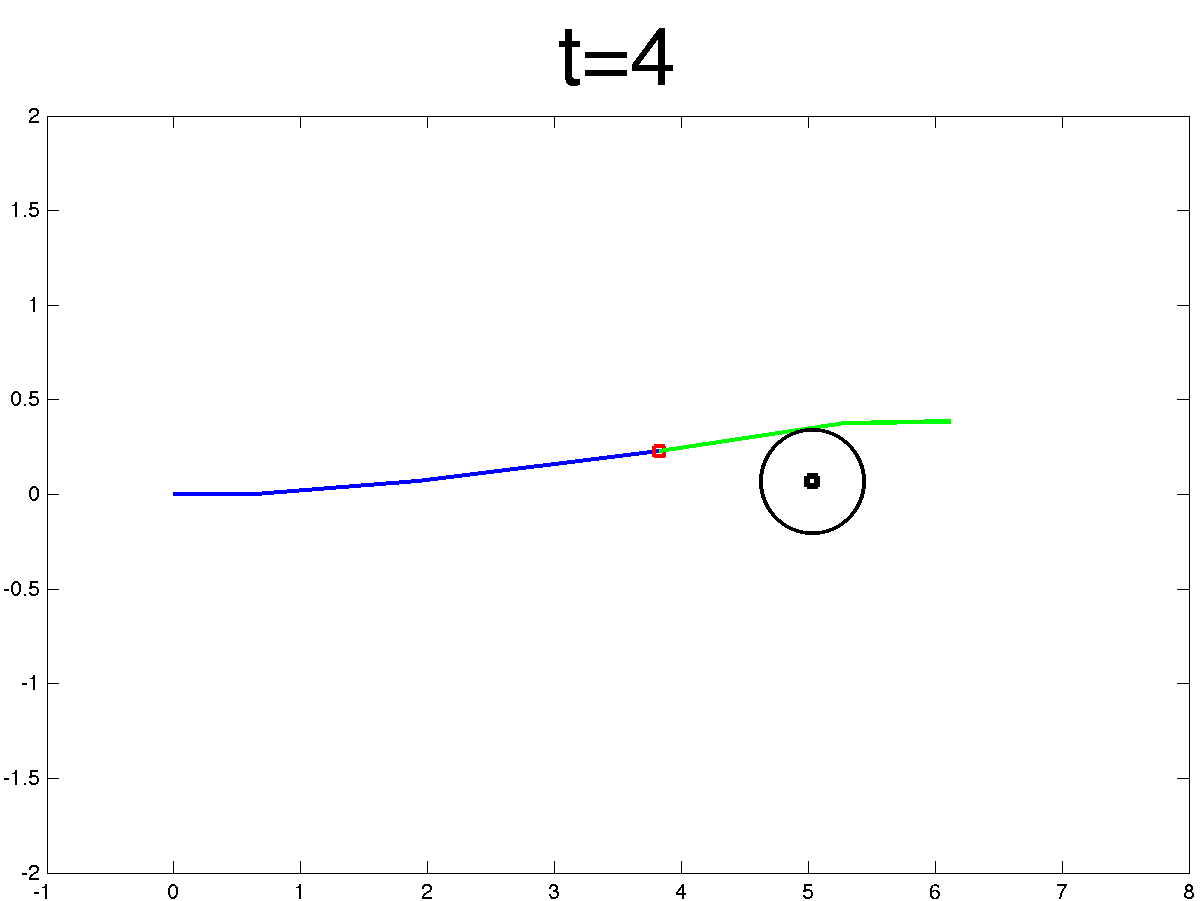} &
\includegraphics[scale=0.13]{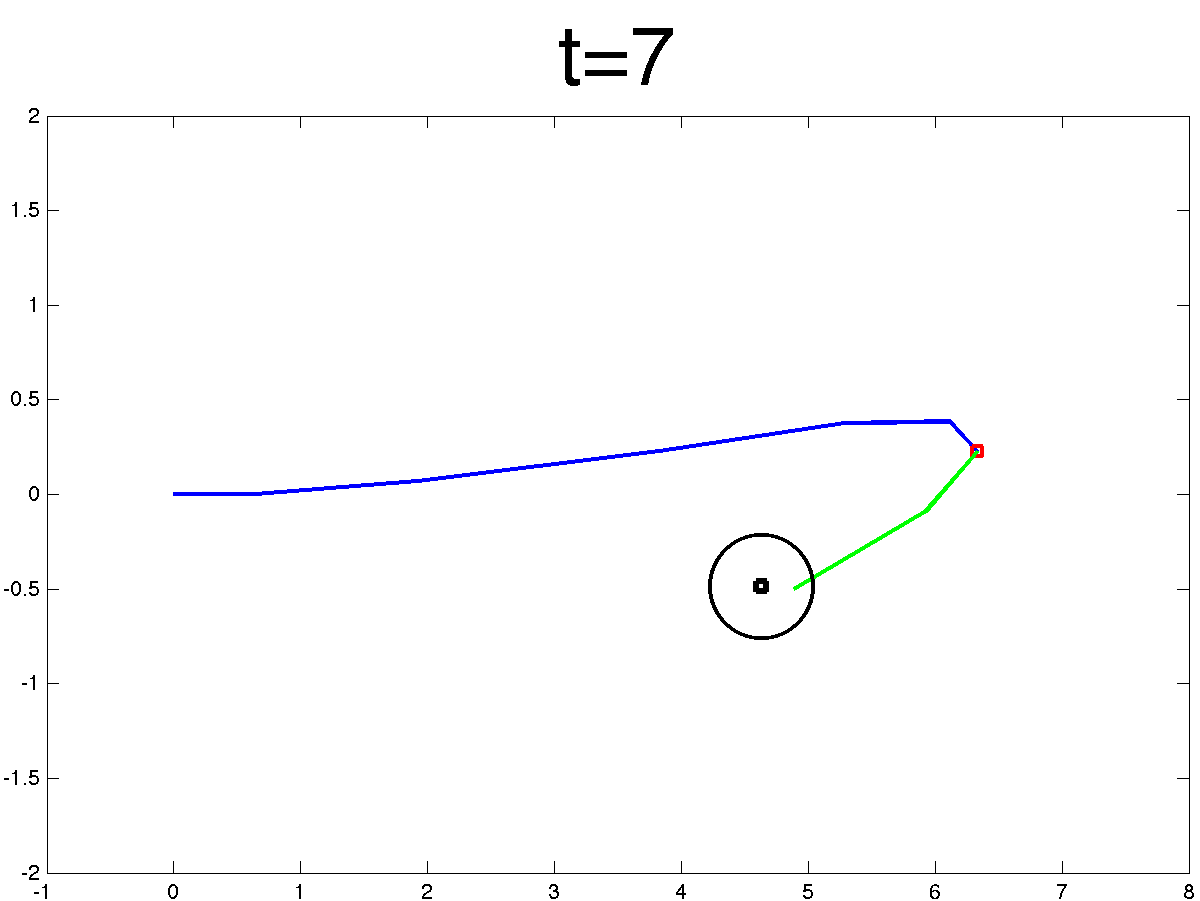} &
\includegraphics[scale=0.13]{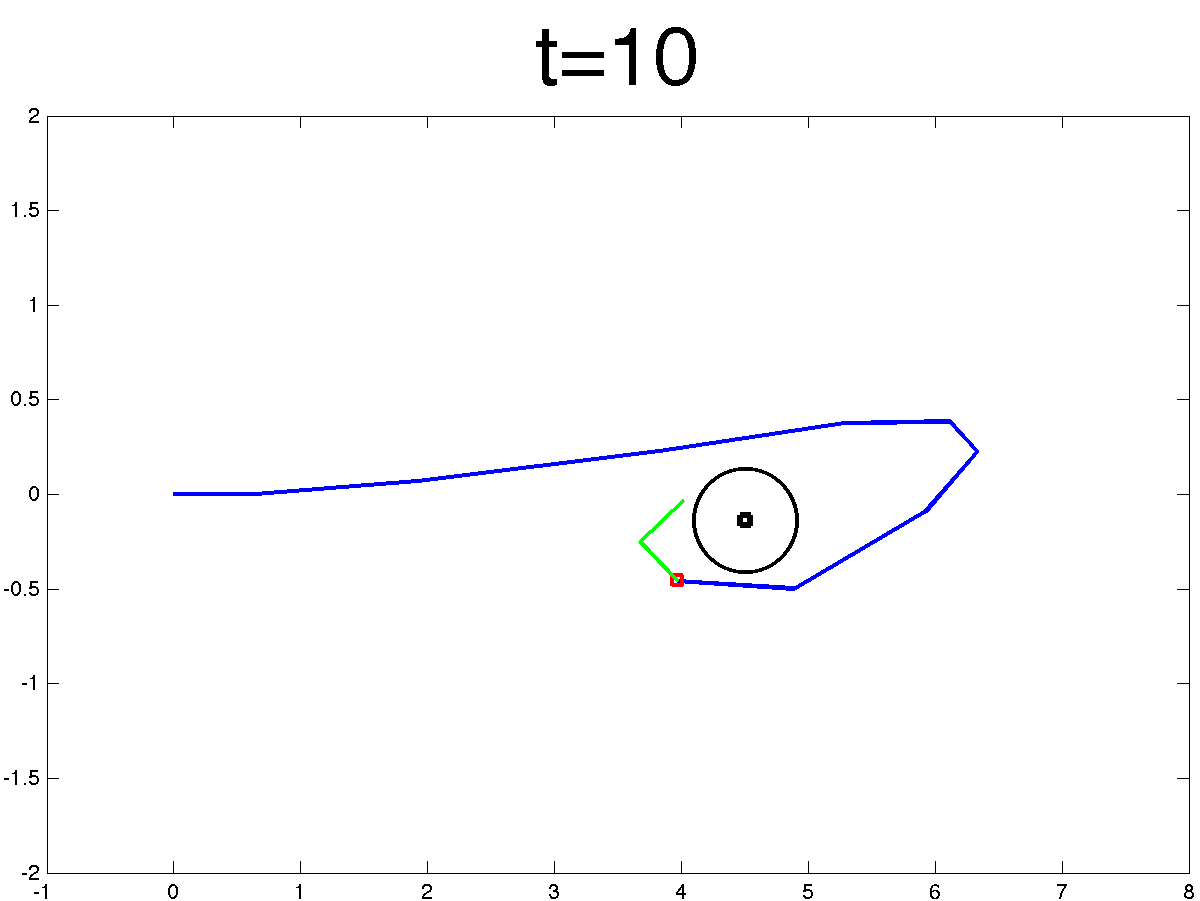} \\

\includegraphics[scale=0.13]{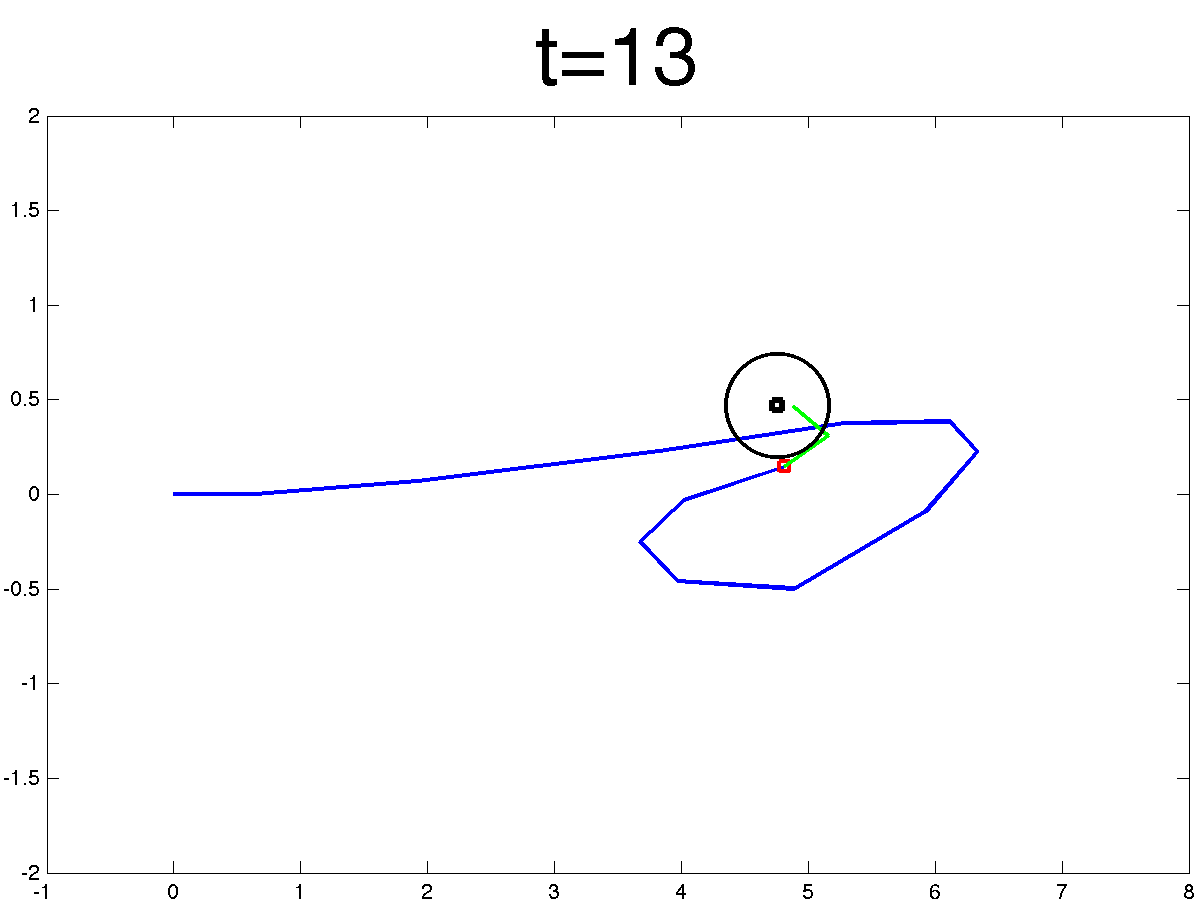} &
\includegraphics[scale=0.13]{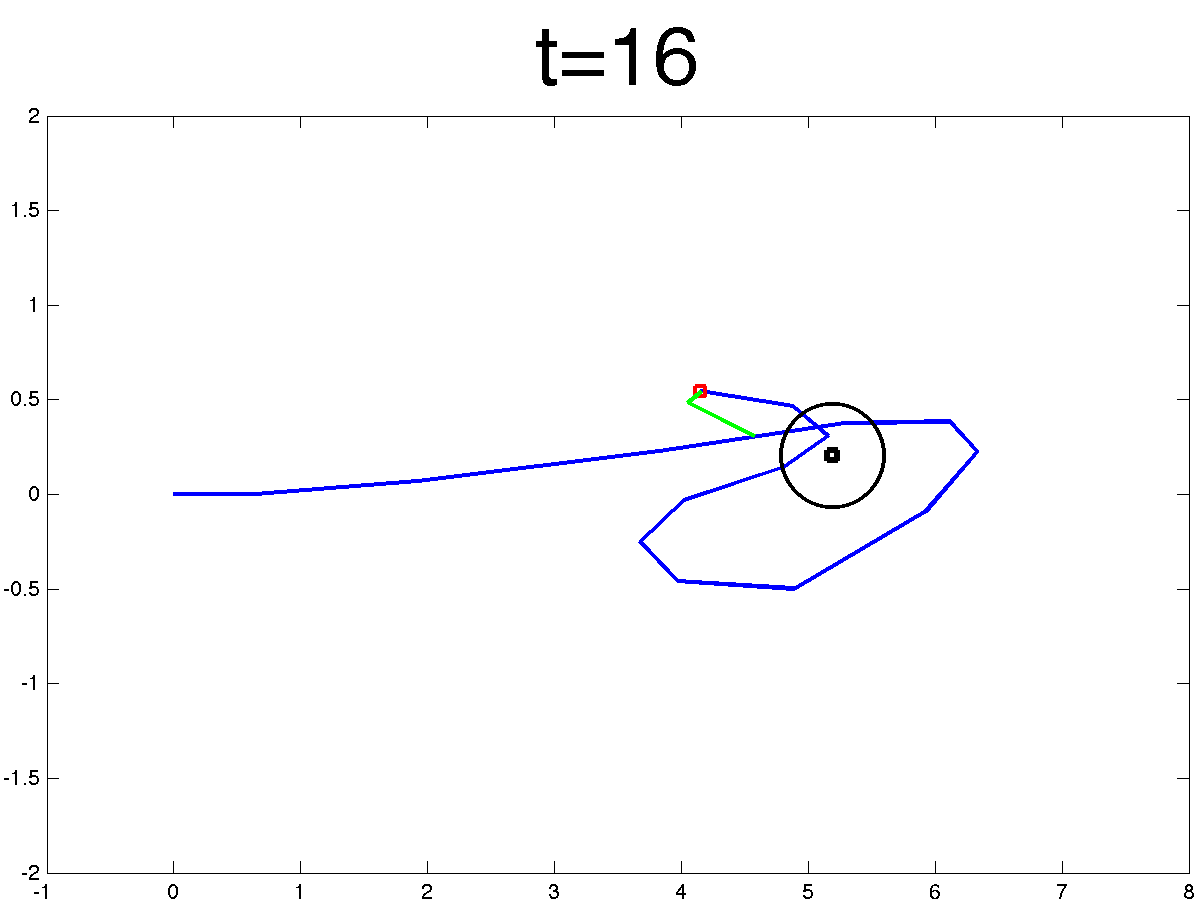} &
\includegraphics[scale=0.13]{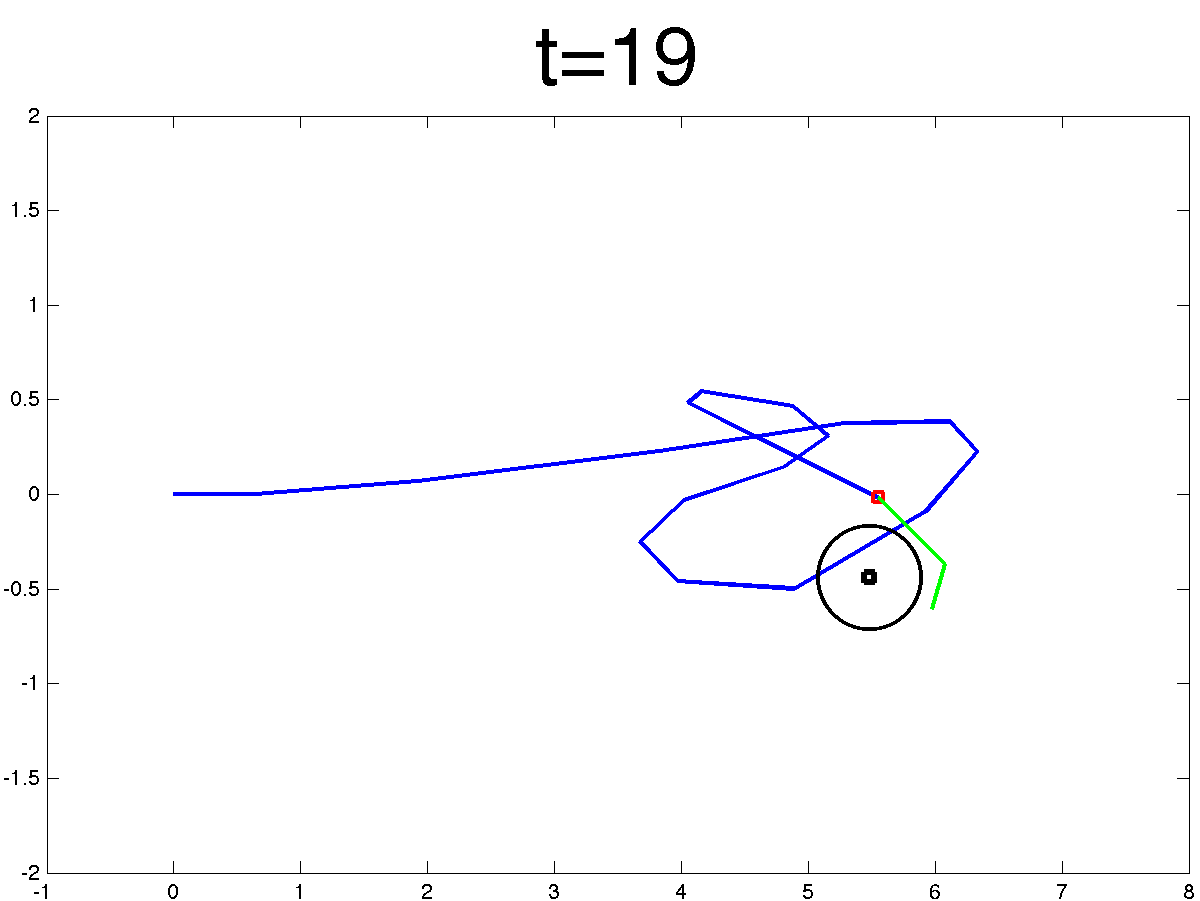} \\

\includegraphics[scale=0.13]{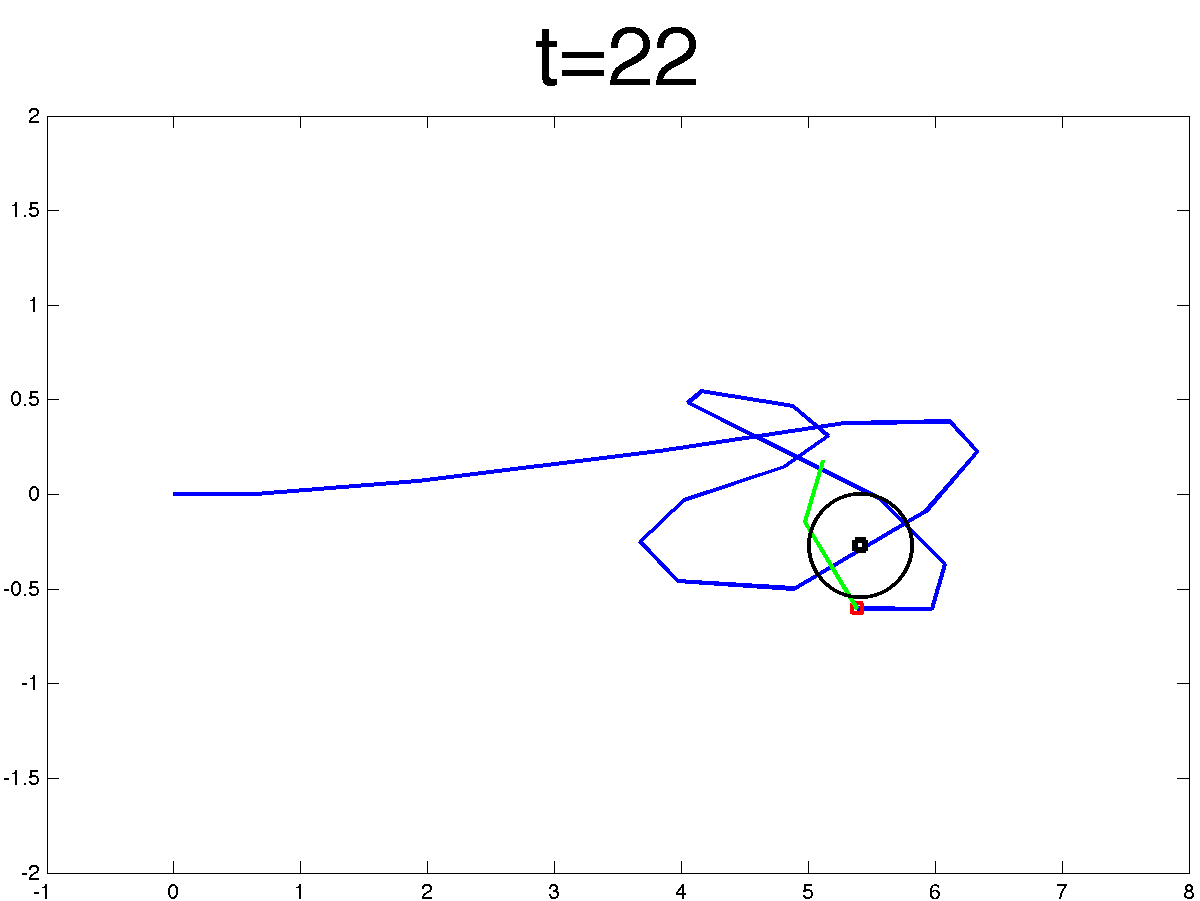} &
\includegraphics[scale=0.13]{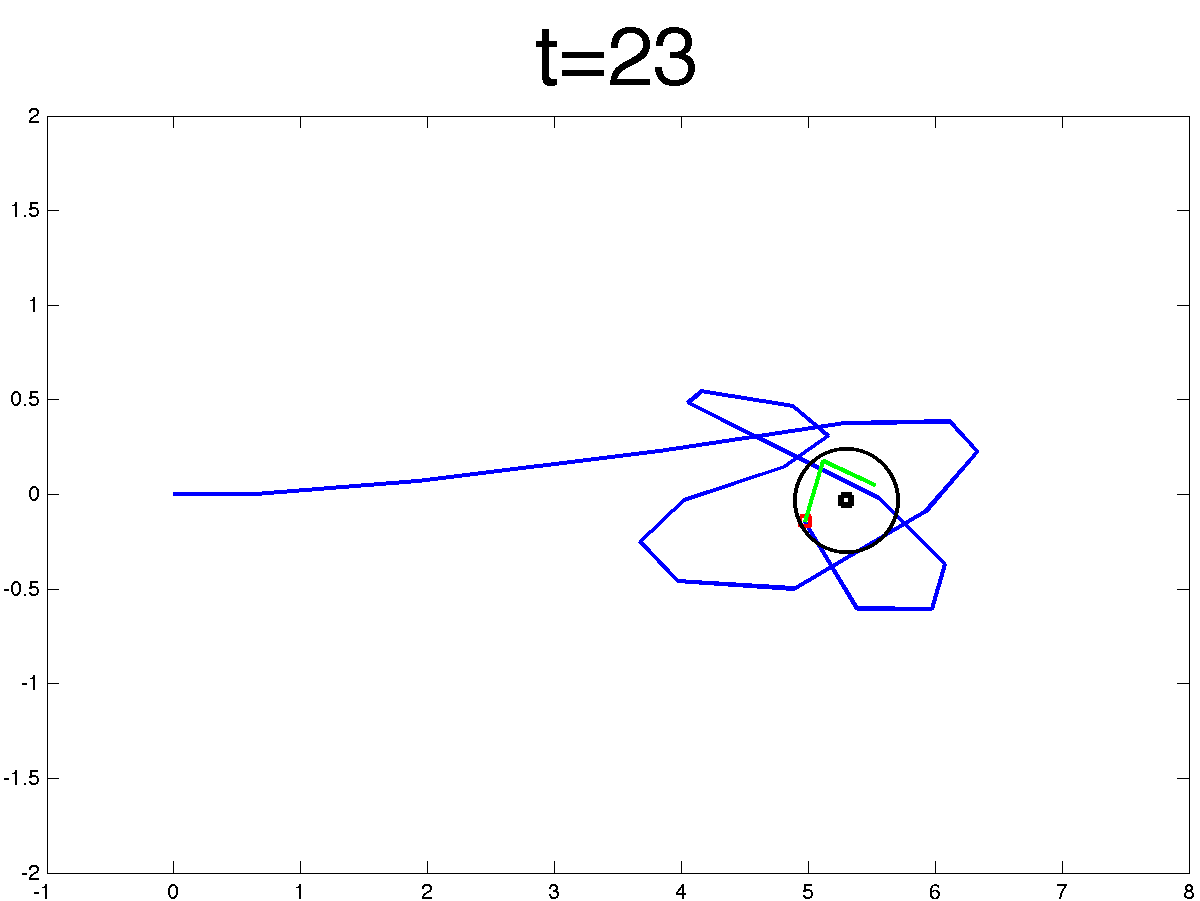} &
\includegraphics[scale=0.13]{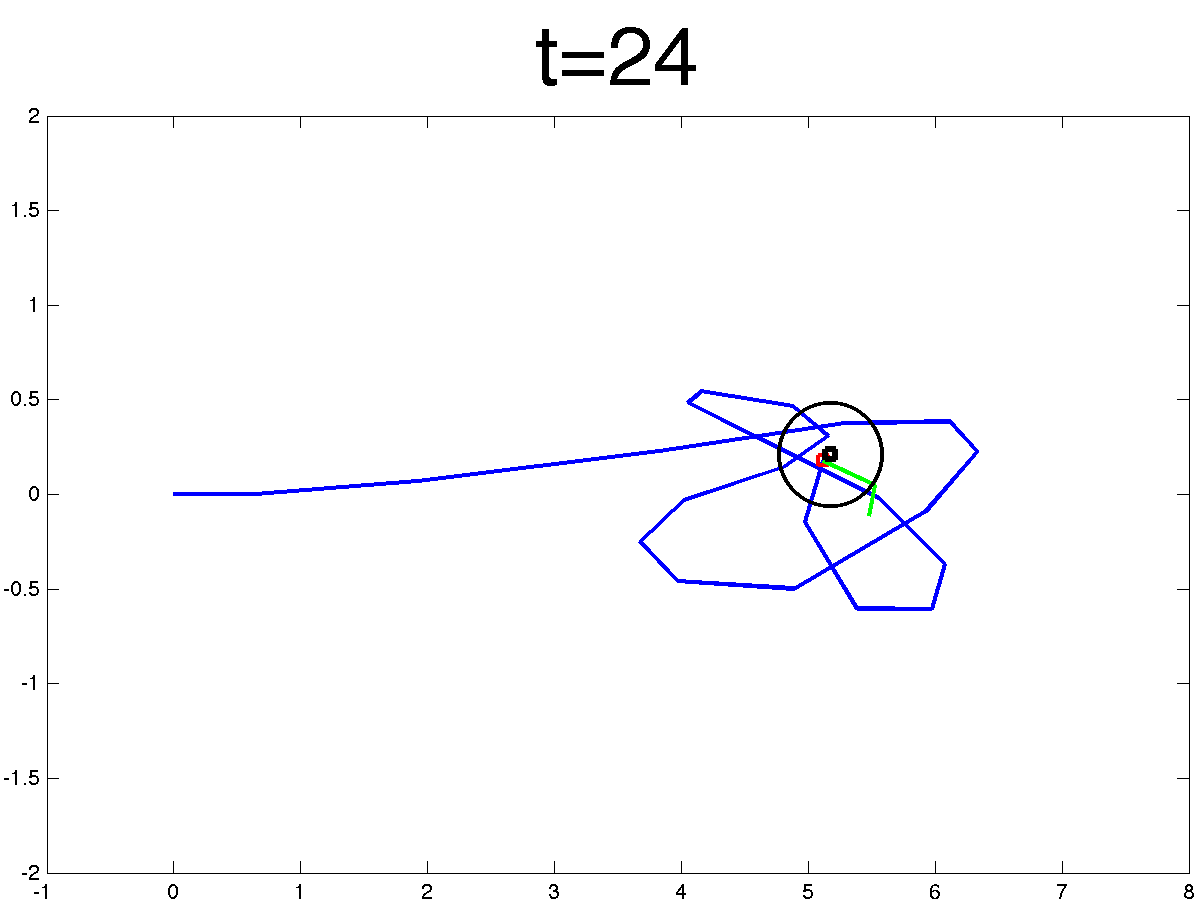} \\

\end{tabular}
\end{centering}
\caption{Example trajectory for the UAV when performing retrieval via interception with rope location. The UAV misses intercepting the center of the rope at $t=10, 13, 16, 22$, but the RHC scheme automatically recovers, until interception is achieved at $t=24$. \label{fig:uav_moving_rope}}
\end{figure}

Since motion planning involves movement in the future, planning  and control techniques need to be robust to uncertainty, such as instrumentation noise, or a changing environment. To address these issues, a  receding time horizon was implemented to solve the UAV recapture  example. In this problem the goal location, a swinging rope, varies with some frequency.  The goal location for the problem is updated to be the current state of the rope, and the MPC problem is solved for five time steps into the future. Of course, if the future location of the rope is known a priori, this problem reduces to those already examined. Resulting trajectories are shown in Figure \ref{fig:uav_moving_rope}.

The capture terminated when the rope was intercepted at $t=24$. Since the trajectory was only planned five time steps ahead, the UAV could not be expected to 
reach its final destination in each of its planning steps. Instead of 
constraining the final position of the UAV to the position of the rope, 
a quadratic penalty, i.e. of the form (\ref{eq:J_cost}), was put on the end condition.  As the condition $\det(R)=0$ corresponds to the UAV stopping altogether,  physically impossible for this application, we restrict $\det(R) > 0.3$ using a square constraint, as in Equation~\ref{eq:det_cons}. Each receding horizon iteration took approximately $t=0.16s$ to compute.

The rope oscillated around the point $(x,y)=(5,0)$ with amplitude along each coordinate of $A_x=0.5$, $A_y=0.5$. The resulting trajectory is shown in Figure \ref{fig:uav_moving_rope}. As can be seen, the receding horizon framework allows for automated planning in the event the rope was missed.


\subsection{Spacecraft}
\begin{figure}
\begin{centering}
\includegraphics[scale=0.22]{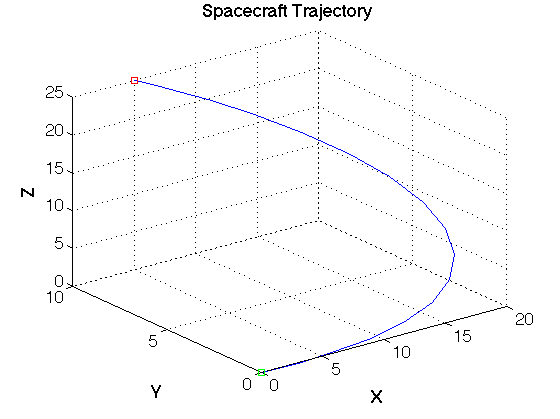}
\includegraphics[scale=0.20]{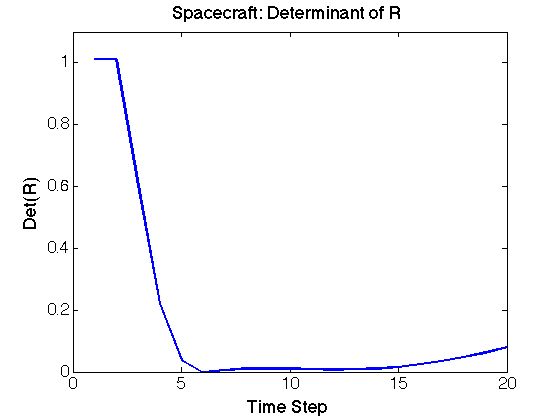}
\end{centering}
\caption{Trajectory for spacecraft maneuever is shown on the left. On the right is the determinant of the rotational state over the course of the maneuver. \label{fig:space_path}}
\end{figure}

\begin{figure}
\begin{centering}
\begin{tabular}{c c}
\includegraphics[scale=0.20]{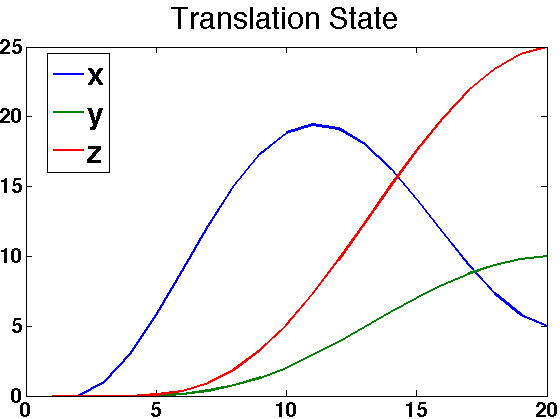} &
\includegraphics[scale=0.20]{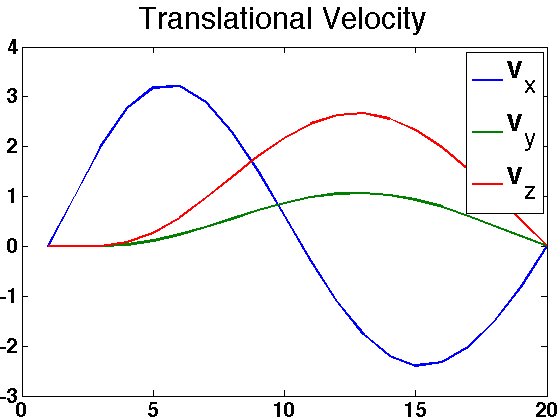} \\

\includegraphics[scale=0.20]{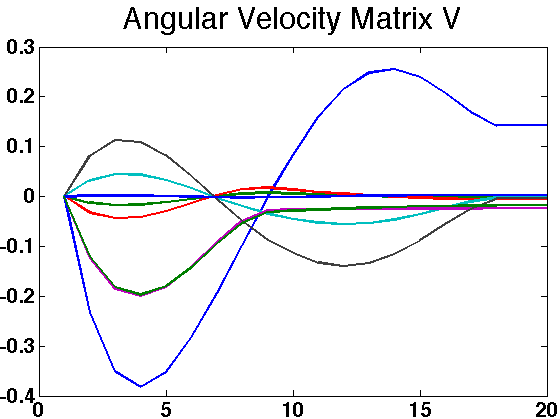} &
\includegraphics[scale=0.20]{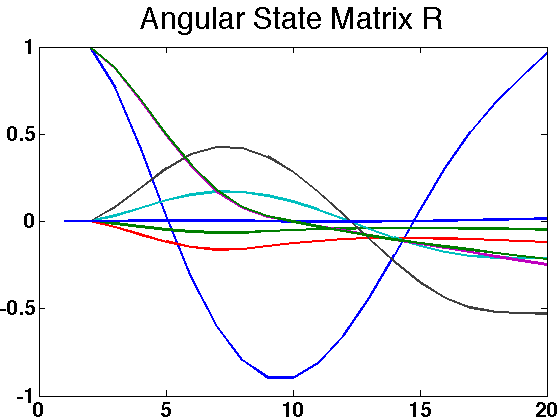} \\

\end{tabular}
\end{centering}
\caption{Trajectory data for the spacecraft maneuver. \label{fig:space_traj}}
\end{figure}

The simple motion planning problem of transfer orbits was used as a 
starting point for the convex optimization experiments and simulations. 
In this problem, a spacecraft containing only simple thrusters moves from one orbit to another. The
spacecraft was modeled as a second-order system in $SE(3)$. This problem required a double integrator
approach because user input controls the acceleration of the spacecraft. 

\begin{eqnarray}
\min &  & \sum_{t=1}^{T}l\left(s\left(t\right),u\left(t\right)\right)+\phi\left(x\left(T\right)\right)\\
s.t. &  & W(t+1)=W(t) + h \cdot u(t) \nonumber \\
      &  & R(t+1)=R(t)+h \cdot W(t) \nonumber \\
      &  & p(t+1)=p(t) + h \cdot R(t)\cdot V \nonumber \\
      &  & s(t+1) = s(t) + h \cdot p(t) \nonumber \\
      &  & R(t)\in SO(n) \nonumber \label{eq:mpc_space}
\end{eqnarray}
where we have now introduced variables $p$ and $W$ to represent the translational and angular velocity, respectively, and
\begin{equation}
V = \left( \begin{array}{c} 1\\ 0 \\ 0 \end{array} \right).
\end{equation}
is a thrust along the $x$ component in the local coordinates of the spacecraft. The spacecraft was given an orientation aligning its thrust with the negative $x$-axis, and a desired end location at $(x,y,z)=(5,10,25)$ that must be achieved with zero velocity.  The resulting trajectory is shown in Figures \ref{fig:space_path} and \ref{fig:space_traj}. As is seen, the determinant drops in the middle of the trajectory, largely allowing the spacecraft to  conserve fuel.

\begin{rem}
Note that current commercial solvers do not support Mixed Integer Semidefinite Programs, precluding the ability to incorporate obstacles or minimum speed constraints for states that evolve on $SO(3)$. However, due to the growth in interest in semidefinite programs, such functionality is anticipated in the near future.
\end{rem}
\section{Discussion}
In this work a novel method to plan over systems with states in $SO(2)$ and $SO(3)$ has been proposed. The method relies on the convex hull of the orbitopes of these groups. The results are convex programs that dispense with transcendental equations and may be solved to obtain a globally optimal solution. The interpretation of the convex hull of these rigid body transformations is shown to correspond to a decrease in speed for our MPC formulation. Furthermore, the presence of these orbitopes is shown not to affect the mixed integer and receding horizon variants of MPC, allowing for obstacles, a minimum speed, and a degree of robustness to be added to the formulation. Examples ranging from the Dubins car to a spacecraft with integrated dynamics has shown the success and wide applicability of the method.

\bibliographystyle{abbrv}
\bibliography{orbitopes,se3_vision}

\end{document}